# DESENVOLVIMENTO DE FERRAMENTA DE SIMULAÇÃO PARA AUXÍLIO NO ENSINO DA DISCIPLINA DE ROBÓTICA INDUSTRIAL

*Afonso Henriques Fontes Neto Segundo – afonsof@unifor.br*
*Professor na Universidade de Fortaleza (UNIFOR)*
*Endereço: Av. Washington Soares, 1321 – Edison Queiroz, Fortaleza – CE, Brasil*

*Joel Sotero da Cunha Neto – joelsotero@unifor.br*
*Professor na Universidade de Fortaleza (UNIFOR)*
*Endereço: Av. Washington Soares, 1321 – Edison Queiroz, Fortaleza – CE, Brasil*

*Halisson Alves de Oliveira – halisson@unifor.br*
*Professor na Universidade de Fortaleza (UNIFOR)*
*Endereço: Av. Washington Soares, 1321 – Edison Queiroz, Fortaleza – CE, Brasil*

*Átila Girão de Oliveira – atilagirao@unifor.br*
*Professor na Universidade de Fortaleza (UNIFOR)*
*Endereço: Av. Washington Soares, 1321 – Edison Queiroz, Fortaleza – CE, Brasil*

*Reginaldo Florencio da Silva – reginaldoflorencio@edu.unifor.br*
*Pesquisador na Universidade de Fortaleza (UNIFOR)*
*Endereço: Av. Washington Soares, 1321 – Edison Queiroz, Fortaleza – CE, Brasil*

**Resumo:** Atualmente a robótica é uma das áreas que mais crescem não apenas no setor industrial, mas também nos setores de consumo e serviços. Diversas áreas se beneficiam com o avanço tecnológico da robótica, como a área militar, a área da saúde, a agricultura, entre outras. A indústria beneficia se com os ganhos em produtividade e qualidade. Porém, para atender a esta crescente demanda, é necessário formar profissionais que já saiam de seus respectivos cursos com um conhecimento mais aprofundado de como projetar e controlar um manipulador robótico. É lógico que para obter esse conhecimento mais aprofundado em robótica, faz se necessário ter uma experiência com um manipulador robótico real, pois a prática tem um poder de fixação muito mais poderoso do que a teoria. Porém, sabe se que um braço robótico não é um investimento baixo, e sua manutenção também não são baratos. Por isso, muitas instituições de ensino não têm condições de fornecer este tipo de experiência a seus alunos. Pensando nisso, através do uso do Unity 3D, que é um software voltado para o desenvolvimento de jogos, foi desenvolvido um simulador de braços robóticos para correlacionar a teoria da sala de aula com o que realmente acontece na prática. Os manipuladores robóticos implementados nesse simulador, podem ser controlados tanto pela cinemática inversa, (que é o padrão usado na indústria), quanto pela cinemática direta.

**Palavras-chave:** Unity 3D. Simulação. Cinemática Inversa. Cinemática Direta. Robótica.

# 1 INTRODUÇÃO

O primeiro braço robótico foi projetado em 1954 por George Devol e construído em 1959 por Devol e Joseph Engelberger. Seu primeiro protótipo chamava se Unimat #001, onde sua primeira aplicação foi na linha de montagem da planta de fundição da General Motors. O sucesso foi tamanho que em 1961 o Unimat 1900 Series tornou se o primeiro robô a ser produzido em massa.(ASSOCIATION, 2018)

Desde então, a robótica é uma área que só tem crescido e não apenas no setor da indústria, ramificando-se para outros setores como a medicina, exploração espacial, agricultura, setor militar e principalmente nos setores de serviços e consumo (WOLFGANG et al., 2017), substituindo a mão de obra em muitos setores, auxiliando seres humanos em outros setores como a medicina e até mesmo substituindo membros de seres humanos por membros biônicos. (POLLOCK, 2015)

Pensando nisto, o intuito deste trabalho é auxiliar os professores e facilitar o ensino da robótica para os alunos que desejam seguir nessa área e muitas vezes não tem acesso a um manipulador robótico para lhe auxiliar no aprendizado.

Este trabalho tem como objetivo desenvolver um simulador para auxiliar o ensino das principais formas de controle de manipuladores robóticos, ajudando o aluno a correlacionar a teoria com a prática através da simulação, amplificando assim seu poder cognitivo de entendimento do conteúdo. Também será possível inserir diferentes modelos de manipuladores robóticos no simulador, permitindo o seu controle pelo usuário, aplicando cinemática inversa e cinemática direta e retornando informações para o usuário sobre o estado atual dos manipuladores.

# 2 ROBÓTICA

O termo robô, deriva da palavra robota, (do idioma tcheco), que significa "trabalhador escravo". Este termo foi utilizado pela primeira vez pelo escritor Tcheco Karel Capek (1890-1938), em uma peça teatral de sua autoria, Rossum's Universal Robots. Essa obra dramatúrgica de 1921 retratava criaturas mecânicas, parecidas com seres humanos, e que eram escravos.

Já o termo robótica, foi utilizado pela primeira vez pelo escritor Asimov, para nomear a ciência que estuda os robôs. Com o avanço da robótica, novas tecnologias foram desenvolvidas e aplicadas, permitindo que os mesmos substituíssem seres humanos com ganhos incríveis em produtividade, qualidade e eficiência. Isso acabou por melhorar a competitividade, aumentar as margens de lucro e saciar a crescente demanda por produtos industrializados.

Hoje em dia existe uma constante busca pelo desenvolvimento de máquinas robóticas cada vez mais sofisticadas, produtivas e capazes de substituir o ser humano em situações de insalubridade, risco de vida ou simplesmente em tarefas repetitivas e entediantes. E apesar dos sindicatos trabalhistas se manifestarem contrários à implementação de robôs na indústria, isto tem se tornado uma prática cada vez mais comum. (SANTOS; JUNIOR, 2015)

## 2.1 Definição de Robô

De acordo com (RIA, 2012), robô industrial é definido como um "manipulador multifuncional reprogramável projetado para movimentar materiais, partes, ferramentas ou peças especiais, através de diversos movimentos programados, para o desempenho de uma variedade de tarefas".

Já a (ISO, 2011), define um robô industrial como sendo: "uma máquina manipuladora com vários graus de liberdade, controlada automaticamente, reprogramável, multifuncional, que pode ter base fixa ou móvel para utilização em aplicações de automação industrial".

Segundo (CRAIG, 2004), manipuladores robóticos são construídos com elos quase rígidos e conectados entre si através de juntas. Essas juntas permitem os movimentos relativos dos elos vizinhos e geralmente são equipadas com sensores de posição, os quais permitem que a posição relativa dos elos vizinhos seja calculada.

## 2.2 Classificação dos manipuladores robóticos

Segundo (ROMANO, 2002), os robôs podem ser classificados de acordo com três diferentes características, são elas: a estrutura mecânica, a geração tecnológica e a participação do operador humano.

A primeira forma de classificar os manipuladores robóticos é de acordo com a estrutura mecânica, onde temos as cinco principais classificações: robô de coordenada cartesiana, Figura 1 (a); robô cilíndrico, Figura 1 (b); robô esférico ou polar, Figura 1 (c); robô SCARA (Selective Compliance Assembly Robot Arm); Figura 1 (d) e robô articulado, Figura 1 (e).

A segunda forma de classificar os manipuladores robóticos é de acordo com a sua geração tecnológica. A primeira geração é denominada sequência fixa, pois só executam a sequência de tarefas definida em sua programação em um ambiente parametrizado (objetos com posicionamentos precisos). Para realizar outras operações eles necessitam ser reprogramados. Os robôs de segunda geração possuem sensores que permitem o robô entender seu ambiente de trabalho e reconhecer as peças as quais devem ser manipuladas por ele, mesmo que elas estejam deslocadas de sua posição ideal. Essa geração é capaz de calcular em tempo real os parâmetros de controle para a realização do movimento. Já a terceira geração é marcada pela intercomunicação entre os robôs e o emprego da inteligência artificial. Os manipuladores dessa geração tem a capacidade de tomar decisões como a rejeição de peças defeituosas e a seleção de uma combinação correta de tolerâncias. (ROMANO, 2002)

Figura 1 – Área de atuação dos robôs

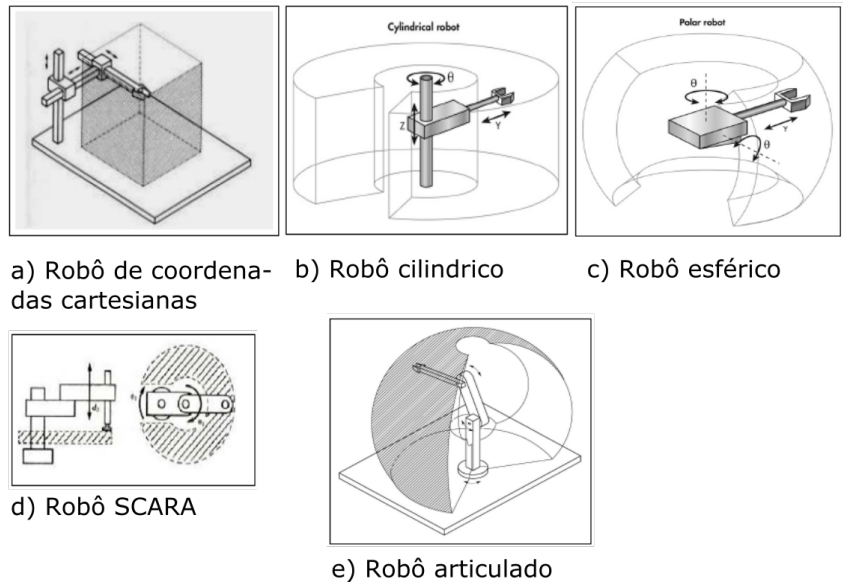

a) Robô de coordenadas cartesianas  b) Robô cilíndrico  c) Robô esférico

d) Robô SCARA

e) Robô articulado

Fonte: (ROMANO, 2002)

A terceira forma de classificar os manipuladores robóticos é de acordo com a participação de um operador humano no processo de controle do robô, sendo determinado pelo quão complexo é o ambiente de trabalho do manipulador, e pela capacidade do manipulador de processar os dados necessários para a execução das tarefas.

Em ambientes estruturados, todos os parâmetros necessários para o bom funcionamento do manipulador podem ser obtidos, e é possível estabelecer um sistema de controle que tenha participação mínima do operador. Neste caso, o sistema é classificado como robótico e são utilizados em atividades como: a soldagem, fixação de circuitos integrados em placas, pintura de superfícies, movimentação de objetos e outros.

Ambientes não estruturados são ambientes que estão em constante mudança, onde não é possível estabelecer parâmetros para que o manipulador possa operar automaticamente, sendo necessário que um operador humano esteja no comando do manipulador para a realização de determinadas tarefas. Este sistema é classificado como tele operado, sendo utilizados em ambientes como: mineração, recuperação de satélites, manipulação de materiais radioativos em usinas ou centros de pesquisas nucleares e exploração de petróleo e gás em plataformas marítimas. (ROMANO, 2002)

## 2.3  Representações espaciais

Segundo (CRAIG, 2004), a manipulação robótica implica dizer que peças e ferramentas serão movimentadas no espaço por algum tipo de mecanismo, criando a necessidade de representar a posição e orientação do manipulador, e dos objetos a serem manipulados. Para representar uma posição no espaço, utiliza se um vetor de posição 3 x 1, Figura 2 (a), e para descrever uma orientação no espaço faz-se necessária a utilização de matrizes de rotação 3 x 3, Figura 2 (b), que nada mais são do que uma forma matemática de representar uma rotação em torno de um eixo de referência. Como pode ser vista na Figura 2 (b), a primeira delas é uma matriz de rotação no eixo z, a segunda matriz representa uma rotação no eixo y e a terceira matriz uma rotação no eixo x.

Figura 2 – Representações espaciais

$$^AP = \begin{bmatrix} p_x \\ p_y \\ p_z \end{bmatrix} \qquad \begin{bmatrix} c\alpha & -s\alpha & 0 \\ s\alpha & c\alpha & 0 \\ 0 & 0 & 1 \end{bmatrix} \begin{bmatrix} c\beta & 0 & s\beta \\ 0 & 1 & 0 \\ -s\beta & 0 & c\beta \end{bmatrix} \begin{bmatrix} 1 & 0 & 0 \\ 0 & c\gamma & -s\gamma \\ 0 & s\gamma & c\gamma \end{bmatrix}$$

a) Vetor de Posição      b) Matriz de Rotação

Fonte: (CRAIG, 2004)

## 2.4  Transformações homogêneas

Ao combinar o vetor posição mostrado na Figura 2 (a), com as matrizes de rotação mostradas na Figura 2 (b), obtém-se uma matriz homogênea de transformação, Figura 3 (a). Segundo (LOPES, 2002), pode-se considerar a matriz de transformação como sendo uma matriz formada por quatro sub matrizes. A sub matriz R3x3, representa as matrizes de rotação da Figura 2 (b), a sub matriz P3x1, representa o vetor posição da Figura 2 (a), a sub matriz f1x3, representa o vetor chamado de efeito de perspectiva, e por ultimo, o quarto elemento da diagonal principal que é denominado de fator de escala. Na robótica, o fator de escala sempre será 1, e o vetor chamado de efeito de perspectiva sempre será preenchido por zeros.

Os três primeiros elementos da quarta coluna da matriz homogênea de transformação, onde está localizada a sub matriz P3x1, representam a aplicação de uma translação em relação a um referencial adotado. Essa translação pode ocorrer ao longo do eixo x, do eixo y, do eixo z, ou ainda, uma combinação dos três. Essa translação é representada de acordo com a matriz que pode ser vista na Figura 3 (b). (LOPES, 2002)

Figura 3 – Matrizes homogêneas

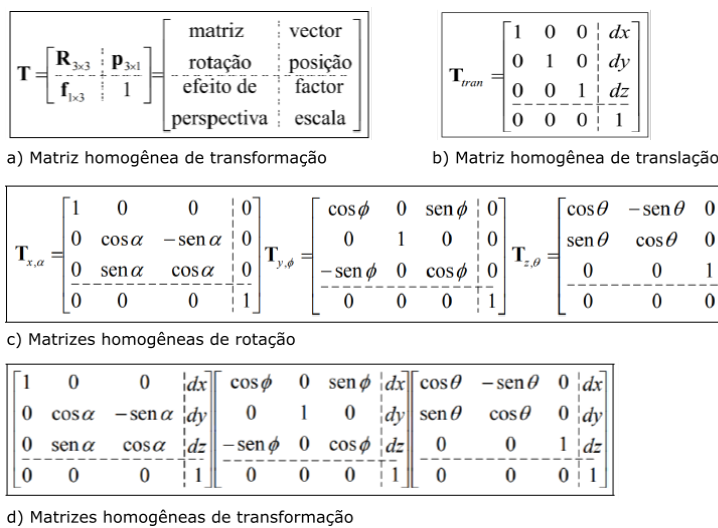

a) Matriz homogênea de transformação
b) Matriz homogênea de translação
c) Matrizes homogêneas de rotação
d) Matrizes homogêneas de transformação

Fonte: (CRAIG, 2004)

Na robótica, caso seja necessário expressar uma operação de rotação, utiliza-se as matrizes homogêneas de rotação, Figura 3 (c). Matrizes homogêneas de rotação nada mais são do que a substituição da sub matriz R3x3 pelas matrizes de rotação apresentadas na Figura 2 (b). Caso seja necessário aplicar rotação e translação simultaneamente, é necessário utilizar as matrizes homogêneas de transformação, Figura 3 (d). Dessa forma, é possível obter a posição e orientação de qualquer objeto em relação a uma referência em um ambiente tridimensional. Além disso, as matrizes homogêneas de transformação são as matrizes utilizadas para se calcular à cinemática direta de manipuladores robóticos.

## 2.5 Cinemática

A cinemática é a matemática por trás da movimentação dos manipuladores robóticos, permitindo que um robô seja capaz de manipular tanto uma ferramenta pesada de usinagem quanto uma frágil peça de isopor. Tudo isso mantendo a perfeita sincronia na rotação de cada uma de suas juntas. (SANTOS; JUNIOR, 2015) O estudo da cinemática nos manipuladores se baseia nas propriedades do movimento que sejam geométricas e baseadas no tempo. (CRAIG, 2004)

Na cinemática dos manipuladores, a pose é o conjunto de coordenadas que as juntas do robô apresentam. Aplicando a geometria nos elos do manipulador é possível elaborar o conjunto de equações que representam a cinemática inversa e direta do mesmo.

- Cinemática direta: é um conjunto de matrizes que ao serem multiplicadas, possibilitam ao usuário encontrar a posição atual da ferramenta. No cálculo da cinemática direta, os ângulos de rotação das juntas e as dimensões dos elos do manipulador já são conhecidos pelo usuário. A única informação que é desconhecida pelo usuário nesta cinemática é a posição final da ferramenta do manipulador.

- Cinemática inversa: é um conjunto de equações que são obtidas após aplicar geometria nos elos do manipulador. Para isso, as dimensões dos elos do manipulador devem ser conhecidas pelo usuário. Na cinemática inversa, o usuário calcula quais devem ser os ângulos das juntas para que a ferramenta do manipulador possa alcançar determinada posição.

É importante notar que a cinemática direta apresenta uma relação única, ou seja, para uma mesma pose, o robô sempre apresentará as mesmas coordenadas de posição. Porém, isto nem sempre ocorre na cinemática inversa e essa característica é conhecida como redundância do robô. (SANTOS; JUNIOR, 2015)

A cinemática inversa baseia-se na geometria dos manipuladores, limitando-se a até três graus de liberdade. Quando aplicada a manipuladores que possuem mais de três graus de liberdade, as juntas excedentes são mantidas sempre fixas. (SANTOS; JUNIOR, 2015)

## 3    SOFTWARES DE DESENVOLVIMENTO

Game engine, (em português, motor de jogo), é o software base para se criar qualquer jogo seja ele no computador, no console, no celular ou tablet. Esse software nada mais é do que um conjunto de bibliotecas capazes de renderizar um jogo em tempo real. (DIAS, 2018)

O software Unity 3D é uma game engine pertencente à empresa Unity Technologies. Sendo um dos vários motores de jogos que existem no mercado, esse software foi escolhido principalmente pela sua grande comunidade, o que possibilita uma curva de aprendizado bem acentuada. Inicialmente é necessário entender a sua interface, onde é possível utilizar diversos layouts de trabalho. A interface de usuário neste trabalho foi desenvolvida pelo autor e é divida em três cenas, são elas:
- Menu, cena que contém todo o menu principal do simulador;
- DirectKinematics, cena que contém toda a cinemática direta;
- InverseKinematics, cena que contém toda a cinemática inversa do simulador.

## 4    DESCRIÇÃO DO PROJETO

Existem vários simuladores educacionais para robótica disponíveis, porém nas pesquisas do autor notou-se uma deficiência quando se trata de robótica industrial, de variedade de manipuladores robóticos ou ainda com relação a sua forma de controle. Como exemplo temos o simulador S-Educ desenvolvido para simular robôs do tipo Lego Mindstorms NXT para alunos entre 7 e 17 anos de idade, ou o simulador VirBot4u que simula apenas um manipulador robótico e não permite que o usuário possa modelar as matrizes da cinemática direta e existe também o RoKiSim que é um simulador educacional multiplataforma que simula o controle de manipuladores reais através da aplicação da cinemática inversa e que apesar de ser extremamente completo, já não é mais suportado pelos seus desenvolvedores desde 2015.

O projeto desenvolvido pelo autor é um software que simula as cinemáticas direta e inversa de diferentes manipuladores robóticos auxiliando o aprendizado na cadeira de robótica industrial. Ao todo foram utilizados oito diferentes modelos de robôs abaixo descritos. Estas variedades de manipuladores ajudam a diversificar as simulações, melhorando o entendimento do aluno.
- WyvernClaws4 foi desenvolvido pelo próprio autor e teve como inspiração diversos manipuladores observados na internet. Figura 4 (a);
- WyvernClaws5 também foi criado pelo autor, feita para ser aplicada a cinemática inversa. Figura 4 (b);

- RMK3 foi obtido através da internet. Figura 4 (c);
- OBJRobot também foi obtido através da internet. Figura 4 (d);
- Robot é outro modelo que foi encontrado na internet. Figura 4 (e);
- SXYxC desenvolvido pela Yamaha, sua nomenclatura completa é SXYxC-D-T1-15-15-ZRSC12-15. Figura 4 (f);
- HXYLx desenvolvido pela Yamaha, e a nomenclatura completa de seu modelo é HXYLx-C-G1-115-25. Figura 4 (g);
- YKX1000 também desenvolvido pela Yamaha pertence à série YKX, sua nomenclatura completa é YKX1000-200 e ele é um manipulador do tipo SCARA. Figura 4 (h).

Figura 4 – Manipuladores

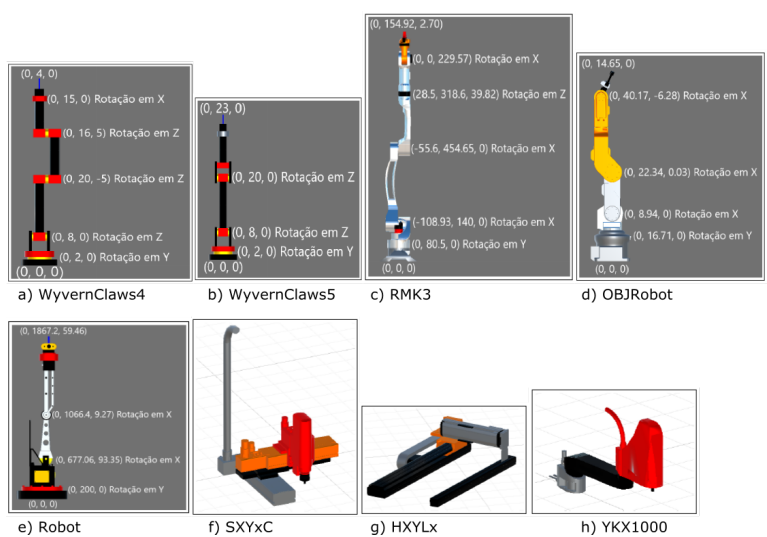

Fonte: Elaborada pelo autor

Para que a cinemática direta possa ser aplicada a cada um dos manipuladores, é necessário primeiramente definir o posicionamento de seus eixos. Isso permitirá que as matrizes de rotação possam ser modeladas. A cinemática direta aplicada nesse simulador é dividida em três partes, são elas:
- Cinemática direta das matrizes identidade;
- Cinemática direta das matrizes pré-definidas;
- Controle dos manipuladores.

A cinemática inversa, como já mencionada anteriormente, é utilizada para determinar o quanto cada junta necessita ser rotacionada para que o manipulador consiga chegar à posição desejada pelo usuário, além disso, ela é limitada a um máximo de três graus de liberdade.

## 5 RESULTADOS

Esse simulador foi desenvolvido com a proposta de ser utilizado por alunos universitários como ferramenta auxiliar para a cadeira de robótica industrial. Com isso em mente, o simulador foi desenvolvido de forma que tivesse os atributos necessários para tornar a sua aplicação valida e útil dentro de uma aula de robótica. Alguns desses atributos é a inserção de diversos manipuladores robóticos, inclusive de diferentes classificações, a possibilidade de permitir que o aluno possa modelar a cinemática direta do manipulador, a inclusão da cinemática inversa, e a presença de modelos reais de manipuladores.

Um questionário com sete perguntas foi aplicado para dez alunos do curso de automação da Universidade de Fortaleza que tiveram a oportunidade de interagir com o simulador. Vale ressaltar que nenhuns dos alunos que participaram do questionário tiveram nenhum tipo de contato anterior com o simulador, e nenhum dos alunos teve nenhum tipo de instrução sobre como utilizar o software. Segui abaixo as questões:

1) Você já cursou a disciplina de robótica industrial?

O intuito principal da pergunta era saber o grau de conhecimento do aluno em relação às cinemáticas de controle. Figura 4 (a).

2) Em uma escala de 1 (muito difícil) a 5 (muito fácil), o quão fácil foi de operar o aplicativo?

A pesquisa mostrou que a interface do software é bem intuitiva e de fácil compreensão. Com poucas melhorias e um melhor refinamento, essa interface pode tornar-se bem fácil de ser compreendida por todos. Figura 4 (b).

3) Em uma escala de 1 (não ajuda) a 5 (ajuda muito), o quanto você acha que uma ferramenta de simulação pode ajudar no ensino da robótica industrial?

O resultado mostrou que os alunos do curso de automação acreditam que um simulador facilitaria e muito o aprendizado da disciplina de robótica industrial. Figura 4 (c).

Figura 4 – Resultado das perguntas do questionário

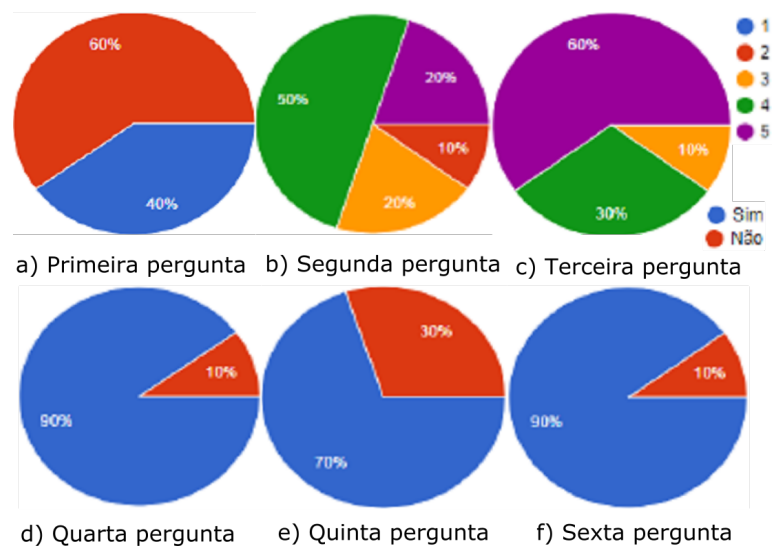

a) Primeira pergunta   b) Segunda pergunta   c) Terceira pergunta

d) Quarta pergunta   e) Quinta pergunta   f) Sexta pergunta

Fonte: Elaborada pelo autor

4) Você considera que sua compreensão geral sobre a cinemática direta de manipuladores robóticos melhorou após interagir com o simulador?

O gráfico com as respostas pode ser visto na Figura 4 (d), e percebe-se que 90% dos alunos responderam que sim, sua compreensão geral sobre a cinemática direta melhorou após utilizar o simulador. Isso mostra o potencial que a utilização de um simulador nas aulas de robótica industrial tem para melhorar e facilitar o ensino dessa disciplina.

5) Você considera que sua compreensão geral sobre cinemática inversa melhorou após interagir com o simulador?

O gráfico obtido pode ser visualizado na Figura 4 (e), e observa-se que 70% dos participantes responderam que sim. Apesar de ser um resultado menor do que os 90% obtidos na pergunta de numero quatro, ainda é um resultado muito expressivo e mostra que há espaço para melhorar a cena da cinemática inversa.

6) De forma geral, você considerou sua interação com o aplicativo útil para o seu aprendizado?

O resultado pode ser visto na Figura 4 (f). Nota-se que 90% dos participantes responderam que a utilização do simulador foi útil para o seu aprendizado. Isso mostra o potencial que a implementação de um simulador de robótica tem para melhorar a compreensão dos alunos.

7) Antes de usar o simulador, como você considerava sua experiência com a visualização de desenhos de braços manipuladores em 2D (para resolução de exercícios)?

Essa era uma pergunta com resposta livre, porém, foi recorrente nas respostas a dificuldade de compreender a visualização de braços robóticos em 2D. Isso pode ser facilmente resolvido com a adoção de um simulador em 3D, o que facilitaria a visualização por parte dos alunos da pose do manipulador.

Ao avaliar todas as respostas obtidas através da aplicação desse pequeno questionário, observa-se que a lacuna existente no ensino da robótica industrial pode ser preenchida pela adoção de um simulador de manipuladores robóticos. Além de ser uma ferramenta que é bem vista pelos alunos, ela tem todo um potencial de tornar a disciplina mais interessante, facilitar a compreensão do tema, reduzir os custos das instituições de ensino, e pode ser melhorada pelos próprios alunos e professores de forma a torna-la ainda melhor.

Ainda é necessário aperfeiçoar a interface e melhorar o software em si. Porém, levando em consideração o feedback positivo obtido com a aplicação do questionário, esse simulador tem potencial para tornar-se uma ferramenta auxiliar nas aulas de robótica.

# 6   CONSIDERAÇÕES FINAIS

Muitas universidades não têm condições de comprar e manter um manipulador robótico real, o que faz com que a aula de robótica industrial acabe tornando-se extremamente teórica, sendo uma das mais importantes para engenheiros que almejam seguir na indústria. Implementar simuladores trás diversos benefícios, como a diminuição dos custos, da quebra de materiais, e possibilidade de controlar diversos manipuladores com diferentes métodos de controle aplicados a eles. É possível observar que este software é uma abordagem totalmente viável e que pode ser implementado em qualquer curso de robótica como um método auxiliar para o ensino do assunto.

Utilizando o Unity 3D, foi possível desenvolver uma interface de usuário bem intuitiva, modelar braços robóticos, incluir manipuladores robóticos já existentes ao software, e aplicar as cinemáticas de controle aos manipuladores e por este ser um trabalho que é voltado para o âmbito educacional, todo o projeto será disponibilizado no GitHub, para que outros desenvolvedores interessados possam contribuir no seu desenvolvimento. O intuito, é transformar esse simulador em uma ferramenta auxiliar, que possa de fato contribuir para a formação de profissionais que já saiam de seus cursos comum a boa compreensão na construção e controle de manipuladores robóticos.

O projeto pode ser encontrado no GitHub através do link: https://github.com/daltonbezerra/Robotic-Arm-with-Unity-3D

**REFERÊNCIAS**

ASSOCIATION, R. I. UNIMATE. Disponível em:

https://www.robotics.org/joseph-engelberger/unimate.cfm. Acesso em: 04 out. 2018.

CRAIG, J. J. **Introduction to Robotics: Mechanics and Control.** [S.l.]: Pearson, 2004. v. 3.

DIAS, R. **Game Engine: o que é, para que serve e como escolher a sua**. Disponível em: https://producaodejogos.com/game-engine/. Acesso em: 04 out. 2018.

INTERNATIONAL ORGANIZATION FOR STANDARDIZATION. **ISO 10218-1:2011:** Robots and robotic devices – safety requirements for industrial robots – part 1: Robots. Geneva, 2011. 43 p.

LOPES, A. M. **Modelação cinemática e dinâmica de manipuladores de estrutura em série.** 101 p. Monografia (Mestrado) — Faculdade de Engenharia, Universidade do Porto, Porto, 2002.

POLLOCK, A. **Pessoas amputadas podem controlar membros biônicos com seus pensamentos.** Disponível em: http://www.administradores.com.br/noticias/tecnologia/pessoas-ambutadas-podem-controlar-membros-bionicos-com-seus-pensamentos/101458. Acesso em: 04 out. 2018.

ROBOTICS INDUSTRIES ASSOCIATION. **ANSI/RIA R15.06-2012:** Industrial robots and robot systems- safety requirements. Michigan, 2012. 160 p.

ROMANO, V. F. **Robótica Industrial Aplicação na Indústria de Manufatura e de Processos.** [S.l.]: Blucher, 2002. v. 1.

SANTOS, W. E. dos; JUNIOR, J. H. G. **ROBÓTICA INDUSTRIAL - FUNDAMENTOS, TECNOLOGIAS, PROGRAMAÇÃO E SIMULAÇÃO**. [S.l.]: Érica, 2015. v. 1.

UNITY. **Relações Publicas**. Disponível em: https://unity3d.com/pt/public-relations. Acesso em: 04 out. 2018.

WOLFGANG, M.; LIKIC, V.; SANDER, A.; KUPPER, **D. Gaining Robotics Advantage. 2017.** Disponível em: https://www.bcg.com/publications/2017/strategy-technology-digital-gaining-robotics-advantage.aspx. Acesso em: 04 out. 2018.

# DEVELOPMENT OF SIMULATION TOOL FOR AID IN THE TEACHING OF INDUSTRIAL ROBOTICS DISCIPLINE

***Abstract:*** *Currently, robotics is one of the fastest growing areas not only in the industrial sector but also in the consumer and service sectors. Several areas benefit from the technological advancement of robotics, especially the industrial area those benefits from gains in productivity and quality. However, to supply this growing demand it is necessary for the newly graduated professionals to have a deeper understanding of how to design and control a robotic manipulator. It is logical that in order to obtain this more in-depth knowledge of robotics, it is necessary to have an experience with a real robotic manipulator, since the practice is a much more efficient way of learning than theory. However, it is known* *that a robotic arm is not a cheap investment, and its maintenance is not cheap either. Therefore, many educational institutions are not able to provide this type of experience to their students. With this in mind, and through the use of Unity 3D, which is a game development software, a robotic arm simulator has been developed to correlate classroom theory with what actually happens in practice. The robotic manipulators implemented on this simulator can be controlled by both inverse kinematics (which is the industry standard) and direct kinematics.*

**Key-words:** *Unity 3D. Simulation. Inverse kinematics. Forward kinematics.*